\pdfoutput=1 % ensure pdflatex (for arXiv)

\documentclass[11pt,a4paper]{article}
\usepackage[hyperref]{acl2021}
\usepackage{times}
\usepackage{latexsym}

\usepackage{nert}
\usepackage{amsmath}
\usepackage{algorithm}
\usepackage[noend]{algpseudocode}
\usepackage{pifont}

\newcommand{\hide}[1]{}

\aclfinalcopy % Uncomment this line for the final submission
\newcommand{\amr}[1]{\texttt{#1}}
\newcommand{\topic}[1]{\paragraph*{#1.}}
\newcommand{\reenttype}[1]{\textsc{#1}\xspace}

\definecolor{calmblue}{RGB}{107,115,213}
\definecolor{dkred}{RGB}{202,0,44} 
\definecolor{dkyellow}{RGB}{231,189,66} 

\newcommand{\abs}[1]{\lvert#1\rvert}
\newcommand{\xmark}{\ding{55}}
\newcommand{\blue}[1]{\textcolor{mdblue}{#1}}
\newcommand{\green}[1]{\textcolor{mdgreen}{#1}}
\newcommand{\red}[1]{\textcolor{orange}{#1}}
\newcommand{\purple}[1]{\red{\textbf{#1}}}
\title{Probabilistic, Structure-Aware Algorithms for Improved Variety, Accuracy, and Coverage of AMR Alignments}

\author{
  Austin Blodgett \quad Nathan Schneider\\
  Georgetown University \\
  \{\emldisplay{ajb341@georgetown.edu}{ajb341}, \emldisplay{nathan.schneider@georgetown.edu}{nathan.schneider}\}\texttt{@georgetown.edu}}
\date{}

\begin{document}
\maketitle
\begin{abstract}

We present algorithms for aligning components of Abstract Meaning Representation (AMR) graphs to spans in English sentences.
We leverage unsupervised learning in combination with heuristics, taking the best of both worlds from previous AMR aligners. Our unsupervised models, however, are more sensitive to graph substructures, without requiring a separate syntactic parse.
Our approach covers a wider variety of AMR substructures than previously considered, achieves higher coverage of nodes and edges, and does so with higher accuracy.
We will release our \textsc{leamr} datasets and aligner for use in research on AMR parsing, generation, and evaluation.

% Current AMR alignment systems focus either on fuzzy token matching and hand written rules, such as JAMR, or MT-style statistical alignment, such as ISI. These strategies fail to produce full-coverage alignments, fail to find novel subgraph alignments, and in some cases produce alignments which are disconnected and are thus not plausible semantically. We present an algorithm for producing full-coverage alignments English spans and AMR subgraphs using an unsupervised algorithm that preserves AMR graph structure. 

% The need for the production of unsupervised AMR alignments can be seen as a bottleneck in performance of AMR parsing as these alignments are typically imperfect and each time the alignments are improved, researchers see a corresponding improvement in parsers trained on the new alignments (Pourdamghani, et al., 2014; Liu, et al., 2018). 

% This alignment algorithm is positioned to improve AMR semantic parsing in future models and do so in a way that preserves interpretability. We also release a dataset of alignments for nodes, relations, and secondary alignments for reentrancies for 60,000 AMRs of English sentences as well as gold alignments for 200 sentences which were hand-aligned for evaluation. 

\end{abstract}

\section{Introduction}
Research with the Abstract Meaning Representation \citep[AMR;][]{amr}, a broad-coverage semantic annotation framework in which sentences are paired with directed acyclic graphs, must contend with the lack of gold-standard alignments between words and semantic units in the English data.
A variety of rule-based and statistical algorithms have sought to fill this void, 
with improvements in alignment accuracy often translating into improvements in AMR parsing accuracy \citep{pourdamghani2014,naseem2019,tamr}. 
%though some algorithms turn instead to latent alignments \citep[e.g.,][]{lyu-18,cai-20}.
Yet current alignment algorithms still suffer from limited coverage and less-than-ideal accuracy, constraining the design and accuracy of parsing algorithms.
%The need for the production of unsupervised alignments can be seen as a bottleneck in performance of AMR parsing and generation as these alignments are typically imperfect and each time the alignments are improved, researchers see a corresponding improvement in parsers trained on the new alignments \cite{pourdamghani2014, naseem2019, tamr}. 
%
%Some recent AMR parsers use latent rather than explicit alignments\nss{cite?}, but 
%Even research on parsers that use latent alignments \citep[e.g.,][]{lyu-18,cai-20} stands to benefit from accurate, human-interpretable alignments in evaluation and error analysis. 
Where parsers use latent alignments \citep[e.g.,][]{lyu-18,cai-20}, explicit alignments can still facilitate evaluation and error analysis.
Moreover, AMR-to-text generation research and applications using AMR stand to benefit from accurate, human-interpretable alignments.
% and error analysis. 
% Aligners are used in research on generation and applications with AMR.

We present \textbf{L}inguistically \textbf{E}nriched \textbf{AMR} (\textsc{leamr}) alignment, %\footnote{Pronounced like ``lemur''.} 
%an unsupervised  alignment algorithm which we use to develop 
which achieves full graph coverage via four distinct types of aligned structures: subgraphs, relations, reentrancies, and duplicate subgraphs arising from ellipsis. 
This formulation lends itself to unsupervised learning of alignment models.
Advantages of our algorithm and released alignments include: (1)~much improved coverage over previous datasets, (2)~increased variety of the substructures aligned, including alignments for all relations, and alignments for diagnosing reentrancies, (3)~alignments are made between spans and connected substructures of an AMR, (4)~broader identification of spans including named entities and verbal and prepositional multiword expressions.

Contributions are as follows:
\begin{itemize}
    \item A novel \textit{all-inclusive} formulation of AMR alignment in terms of mappings between spans and connected subgraphs, including spans aligned to multiple subgraphs; mappings between spans and inter-subgraph edges; and characterization of reentrancies. Together these alignments fully cover the nodes and edges of the AMR graph (\cref{sec:formulation}).
    \item An algorithm combining rules and EM to align English sentences to AMRs without supervision (\cref{sec:aligner}), achieving higher coverage and quality than existing AMR aligners (\cref{sec:results}).
    \item A corpus with automatic alignments for LDC2020 and \emph{Little Prince} data as well as a few hundred manually annotated sentences for tuning and evaluation (\cref{sec:data-release}).
\end{itemize}
We release this dataset of alignments for over 60,000 sentences % 60,000 AMRs for English sentences in the LDC2020T02 dataset as well as gold alignments for 200 sentences which were hand-aligned for evaluation. 
along with our aligner code to facilitate more accurate models and greater interpretability in future AMR research.
%These resources are positioned to improve AMR semantic parsing in future models and do so in a way that preserves interpretability. 

\section{Related Work}\label{sec:relwork}
 
The main difficulty presented by AMR alignment is that it is a many-to-many mapping problem, with gold alignments often mapping multiple tokens to multiple nodes while preserving AMR structure. Previous systems use various strategies for aligning. They also have differing approaches to what types of substructures of AMR are aligned---whether they are nodes, subgraphs, or relations---and what they are aligned to---whether individual tokens, token spans, or syntactic parses. Two main alignment strategies remain dominant, though they may be combined or extended in various ways: rule-based strategies as in \citet{jamr}, \citet{jamr2}, \citet{tamr}, and  \citet{szubert2018}, and statistical strategies using Expectation-Maximization as in \citet{pourdamghani2014}.

\topic{JAMR} The JAMR system \citep{jamr,jamr2} aligns token spans to subgraphs using iterative application of an ordered list of 14 rules which include exact and fuzzy matching. %and take advantage of idiosyncrasies of AMR notation (e.g., the particular structures of governments, countries, dates, etc.). 
JAMR alignments form a connected subgraph of the AMR by the nature of the rules being applied. A disadvantage of JAMR is that it lacks a method for resolving ambiguities, such as repeated tokens, or of learning novel alignment patterns.

\topic{ISI} The ISI system \citep{pourdamghani2014} produces alignments between tokens and nodes and between tokens and relations via an Expectation-Maximization (EM) algorithm in the style of IBM Model~2 \citep{brown-88}. First, the AMR is linearized; then EM is applied using a symmetrized scoring function of the form \(P(a\mid t)+P(t\mid a)\), where $a$ is any node or edge in the linearized AMR and $t$ is any token in the sentence. Graph connectedness is not enforced for the elements aligning to a given token.
%ISI alignments may not form connected substructures of the AMR as a token can be aligned to multiple nodes or edges that need not be connected. 
Compared to JAMR, ISI produces more novel alignment patterns, but also struggles with rare strings such as dates and names, where a rule-based approach is more appropriate. 

\topic{Extensions and Combinations} TAMR \citep[Tuned Abstract Meaning Representation;][]{tamr} uses the JAMR alignment rules, along with two others, to produce a set of candidate alignments for the sentence. Then, the alignments are ``tuned'' with a parser oracle to select the candidates that correspond to the oracle parse that is most similar to the gold AMR. 

Some AMR parsers \citep{naseem2019,astudillo2020} use alignments which are a union of alignments produced by the JAMR and ISI systems. The unioned alignments achieve greater coverage, improving parser performance.

\topic{Syntax-based} Several alignment systems attempt to incorporate syntax into AMR alignments. \Citet{chen2017} perform unsupervised EM alignment between AMR nodes and tokens, taking advantage of a Universal Dependencies (UD) syntactic parse as well as named entity and semantic role features. \Citet{szubert2018} and \citet{chu2016} both produce hierachical (nested) alignments between AMR and a syntactic parse. \citeauthor{szubert2018} use a rule-based algorithm to align AMR subgraphs with UD subtrees. \citeauthor{chu2016} use a supervised algorithm to align AMR subgraphs with constituency parse subtrees. 

\topic{Word Embeddings} Additionally, \citet{portugueseAMR} use an alignment method designed to work well in low-resource settings using pretrained word embeddings for tokens and nodes.

\topic{Graph Distance} \citet{wang2017-getting} use an HMM-based aligner to align tokens and nodes. They include in their aligner a calculation of graph distance as a locality constraint on predicted alignments. This is similar to our use of projection distance as described in \cref{sec:aligner}.

\begin{table}[t]
    \centering\small
    \begin{tabular}{c|c|c|c}
        & nodes & edges & reentrancies \\
        JAMR & 91.1 & \xmark & \xmark \\
         ISI & 78.7 & 9.8 & \xmark \\
         TAMR$^*$ & 94.9 & \xmark & \xmark \\
    \end{tabular}
    \caption{Coverage and types of previous alignment systems. Scores are evaluated on 200 gold test sentences. $^*$TAMR is evaluated on a subset of 91 sentences.}
    \label{tab:coverage}
\end{table}

\begin{figure*}
\centering\small
\begin{tabular}{@{}b{19.5em}@{}b{31em}}
    \begin{tabular}{lllll}
        \multicolumn{5}{l}{\amr{(\blue{w} / want-01}}\\
        &\multicolumn{4}{l}{\amr{\purple{:ARG0} (\blue{p} / person}}\\
        &&\multicolumn{3}{l}{\amr{\blue{:ARG0-of} (\blue{s} / study-01)}}\\
        &&\multicolumn{3}{l}{\amr{\red{:ARG1-of} (\blue{i} / include-91}}\\
        &&&\multicolumn{2}{l}{\amr{\red{:ARG2} (\green{p2} / person}}\\
        &&&&\multicolumn{1}{l}{\amr{\green{:ARG0-of} (\green{s2} / study-01))}}\\
        &&&\multicolumn{2}{l}{\amr{\red{:ARG3} (\blue{m} / most)))}}\\
        &\multicolumn{4}{l}{\amr{\red{:ARG1} (\blue{v} / visit-01}}\\
        &&\multicolumn{3}{l}{\amr{\purple{:ARG0 p}}}\\
        &&\multicolumn{3}{l}{\amr{\red{:ARG1} (\blue{c} / city \blue{:name} (\blue{n} / name }}\\
        &&&\multicolumn{2}{l}{\amr{\blue{:op1 "New" :op2 "York"}))}}\\
        &&\multicolumn{3}{l}{\amr{\red{:time} (\blue{g} / graduate-01}}\\
        &&&\multicolumn{2}{l}{\amr{\purple{:ARG0 p})))}}\\
        \\
    \end{tabular} 
%\end{tabular}
& \hphantom{xxx}\newline % necessary for proper alignment of nested table, don't ask me why
    \begin{tabular}{|@{~~}r@{~}l|r@{~}l@{~~}|}
        \hline
        \multicolumn{2}{|c|}{\textbf{Subgraph Alignments}} &                     \multicolumn{2}{c|}{\textbf{Relation Alignments}} \\
        \w{Most} $\rightarrow$ & \blue{\amr{m}}, &        \w{of} $\rightarrow$ & \amr{\red{s :ARG1-of i}}, \\
        \w{of} $\rightarrow$ & \blue{\amr{i}}, &          & \amr{\red{i :ARG2 p2}}, \\
        \color{gray}\w{the} $\rightarrow$ & \color{gray} $\emptyset$, &        & \amr{\red{i :ARG3 m}}; \\
        \w{students} $\rightarrow$ & \amr{(\blue{p :ARG0-of s})}, &\w{want} $\rightarrow$ & \amr{\red{w :ARG0 p}}, \\
        \w{want} $\rightarrow$ & \blue{\amr{w}}, & & \amr{\red{w :ARG1 v}}; \\
        \color{gray}\w{to} $\rightarrow$ & \color{gray} $\emptyset$, &         \w{visit} $\rightarrow$ & \amr{\red{v :ARG0 p}}, \\
        \w{visit} $\rightarrow$ & \blue{\amr{v}}, &        & \amr{\red{v :ARG1 c}}; \\
        \w{New York} $\rightarrow$ & \amr{(\blue{c :name}} &         \w{graduate} $\rightarrow$ & \amr{\red{g :ARG0 p}};\\
        \multicolumn{2}{|r|}{\amr{(\blue{n :op1 "New" :op2 "York"}))},} & \w{when} $\rightarrow$ & \red{\amr{v :time g}} \\
        \cline{3-4}
        \color{gray} \w{when} $\rightarrow$ & \color{gray} $\emptyset$, & \multicolumn{2}{c|}{\textbf{Reentrancy Alignments}} \\  
        \color{gray} \w{they} $\rightarrow$ & \color{gray} $\emptyset$, & \w{want} $\rightarrow$ & \amr{\purple{w :ARG0 p}} (\reenttype{primary}), \\
        \w{graduate} $\rightarrow$ & \blue{\amr{g}} &         & \amr{\purple{v :ARG0 p}} (\reenttype{control}); \\
        \cline{1-2}
        \multicolumn{2}{|c|}{\textbf{Duplicate Subgraphs}} & \w{they} $\rightarrow$ & \amr{\purple{g :ARG0 p}} (\reenttype{coref}) \\
        \w{students} $\rightarrow$ & \amr{(\green{p2 :ARG0-of s2})} & & \\
        \hline
        \end{tabular}
\end{tabular}
    \caption{AMR and alignments for the sentence
     ``\textit{Most of the students want to visit New York when they graduate.}'' Alignments are differentiated by colors: blue (subgraphs), green (duplicate subgraphs), and orange (relations). Relations that also participate in reentrancy alignments are bolded. 
     }
    \label{fig:alignment_layers}
\end{figure*}

\topic{Drawbacks of Current Alignments}
Alignment methods  vary in terms of components of the AMR that are candidates for alignment. Most systems either align nodes (e.g., ISI) or connected subgraphs (e.g., JAMR), with incomplete coverage. Most current systems do not align \emph{relations} to tokens or spans, and those that do (such as ISI) do so with low coverage and performance. None of the current systems align reentrancies, although \citet{szubert-20} developed a rule-based set of heuristics for identifying reentrancy types. \Cref{tab:coverage} summarizes the coverage and variety of prominent alignment systems. 

\section{An All-Inclusive Formulation of AMR Alignment}\label{sec:formulation}

Aligning AMRs to English sentences is a vexing problem not only because the English training data lacks gold alignments, but also because AMRs---unlike many semantic representations---are not \emph{designed} with a derivational process of form--function subunits in mind. Rather, each AMR graph represents the full-sentence meaning, and AMR annotation conventions can be opaque with respect to the words or surface structure of the sentence, e.g., by unifying coreferent mentions and making explicit certain elided or pragmatically inferable concepts and relations. 
Previous efforts toward general tools for AMR alignment have considered mapping tokens, spans, or syntactic units to nodes, edges, or subgraphs (\cref{sec:relwork}).
Other approaches to AMR alignment have targeted specific compositional formalisms \citep{groschwitz2018amr, beschke2019, blodgett2019}.

We advocate here for a definition of alignment that is \emph{principled}---achieving full coverage of the graph structure---while being \emph{framework-neutral} and \emph{easy-to-understand}, by aligning graph substructures to shallow token spans on the form side, rather than using syntactic parses. We do use structural considerations to constrain alignments on the meaning side, but by using spans on the form side, we ensure the definition of the alignment search space is not at the mercy of error-prone parsers.

\topic{Definitions} Given a tokenized sentence $\mathbf{w}$ and its corresponding AMR graph $\mathcal{G}$, a complete alignment 
assumes a segmentation of $\mathbf{w}$ into spans $\mathbf{s}$, each containing one or more contiguous tokens;
and puts each of the nodes and edges of $\mathcal{G}$ in correspondence with some span in $\mathbf{s}$. %\footnote{The approach presented in \cref{sec:algorithm} commits to a span segmentation before predicting alignments.} 
%Additionally, our formulation requires that 
%Every part of the AMR graph is thus aligned to exactly one span. 
A span may be aligned to one or more parts of the AMR, or else is null-aligned.
Individual alignments for a sentence are grouped into four \textbf{layers}: subgraph alignments, duplicate subgraph alignments, relation alignments, and reentrancy alignments. These are given for an example in \cref{fig:alignment_layers}.

All alignments are between a single span and a substructure of the AMR.
A span may be aligned in multiple layers which are designed to capture different information.
Within the subgraph layer, alignments are mutually exclusive with respect to both spans and AMR components. The same holds true within the relation layer.
Every node will be aligned exactly once between the subgraph and duplicate subgraph layers. Every edge will be aligned exactly once between the subgraph and relation layers, and may additionally have a secondary alignment in the reentrancy layer.

\subsection{Subgraph Layer}\label{sec:subgraphs-layer}

Alignments in this layer generally reflect the lexical semantic content of words in terms of connected,\footnote{Nodes aligned to a span must form a connected subgraph  with two exceptions: (1)~duplicate alignments are allowed and are separated into subgraph and duplicate layers; (2)~a span may be aligned to two terminal nodes that have the same parent. For example, \w{never} aligns to \amr{:polarity - :time ever}, two nodes and two edges which share the same parent.} directed acyclic subgraphs of the corresponding AMR. Alignments are mutually exclusive (disjoint) on both the form and meaning sides.

\subsection{Duplicate Subgraph Layer}\label{sec:duplicates-layer}

A span may be aligned to multiple subgraphs if one is a duplicate of the others, with a matching concept. This is often necessary when dealing with ellipsis constructions, where there is more semantic content in the AMR than is pronounced in the sentence and thus several identical parts of the AMR must be aligned to the same span. In this case, a single subgraph is chosen as the primary alignment (whichever is first based on depth-first order) and is aligned in the subgraph alignment layer, and any others are represented in the duplicates alignment layer.
For example, verb phrase ellipsis, as in \pex{I swim and so do you}, would involve duplication of the predicate \w{swim}, with distinct \amr{ARG0}s. Similarly, in \cref{fig:alignment_layers}, \pex{Most of the students} involves a subset-superset structure where the subset and superset correspond to separate nodes. Because \w{student} is represented in AMR like \w{person who studies}, there are two 2-node subgraphs aligned to \w{student}, one with the variables \amr{p} and \amr{s}, and the duplicate with \amr{p2} and \amr{s2}.
\hide{Examples with duplicates include:
\begin{itemize}
    \item \pex{Most of the students} (\cref{fig:alignment_layers}): AMR conventions treat this as a subset-superset structure with \amr{include-91} where the subset and superset correspond to separate nodes. Because the word \w{student} is represented in AMR like \w{person who studies} there are two 2-node subgraphs aligned to \w{student}, one with the variables \amr{p} \& \amr{s}, and the duplicate with \amr{p2} \& \amr{s2}.
    \item Verb phrase ellipsis, as in \pex{I swim and so do you}, would involve duplication of the predicate \w{swim}, with distinct \amr{ARG0}s.
    \item AMR annotators might express pragmatic inferences as concept duplication. 
    For instance, the second part of \pex{She wanted to swim but \textbf{I was undecided}} might be interpreted as `I was undecided \textbf{about swimming}', requiring a second AMR predicate aligned to \w{swim}.
\end{itemize}}
The difficulty that duplicate subgraphs pose for parsing and generation makes it convenient to put these alignments in a separate layer.

\subsection{Relation Layer}\label{sec:relations-layer}

This layer includes alignments between a span and a single relation---such as \w{when} $\rightarrow$ \amr{:time}---and alignments mapping a span to its argument structure---such as \w{give} $\rightarrow$ \amr{:ARG0 :ARG1 :ARG2}. All edges in an AMR that are not contained in a subgraph fit into one of these two categories.

English function words such as prepositions %relativizers, \nss{we don't actually align relativizers}
and subordinators typically function as connectives between two semantically related words or phrases, and can often be identified with the semantics of AMR relations. But many of these function words are highly ambiguous. Relation alignments make their contribution explicit.
For example, \w{when} in \cref{fig:alignment_layers} aligns to a \amr{:time} relation.
% \begin{itemize}
%     \item \pex{Let's talk \textbf{when} the meeting is over} gets an alignment \w{when} $\rightarrow$ \amr{:time}
%     \item \pex{Let's meet \textbf{at} the restaurant} gets an alignment \w{at} $\rightarrow$ \amr{:location}
% \end{itemize}

For spans that are aligned to a subgraph, incoming or outgoing edges attached to that subgraph may also be aligned to the span in the relation layer. These can include core or non-core roles as long as they are evoked by the token span. 
For example, \cref{fig:alignment_layers} contains \w{visit} $\rightarrow$ \amr{:ARG0 :ARG1}. 

\subsection{Reentrancy Layer}\label{sec:reentrancies-layer}

A \textbf{reentrant} node is one with multiple incoming edges. 
In \cref{fig:alignment_layers}, for example, \amr{p} appears three times: 
once as the \amr{ARG0} of \amr{w} (the wanter), 
once as the \amr{ARG0} of \amr{v} (the visitor), 
and once as the \amr{ARG0} of \amr{g} (the graduate).
The \amr{p} node is labeled with the concept \amr{person}---in the PENMAN notation 
used by annotators, each variable's concept is only designated on one occurrence of the variable, the choice of occurrence being, in principle, arbitrary.\hide{\footnote{Our aligner reads the annotations into a graph data structure that merges all occurrences of the variable into one node.}}
These three \amr{ARG0} relations are aligned to their respective predicates in the relation layer.
But there are many different causes of reentrancy, and AMR parsers stand to benefit from additional information about the nature of each reentrant edge,
such as the fact that the pronoun \w{they} is associated with one of the \amr{ARG0} relations.

\begin{table*}[h!]
\centering\small
\begin{tabular}{r>{\raggedright}p{18em}@{~~~}>{\pex\bgroup}p{14.5em}<{\egroup}}
\textbf{Type} & \textbf{Triggered by} & \textbf{Example} \\
\hline
\reenttype{coref} & a pronoun (including possessive or reflexive) (anaphora) & I love \uline{my} \textbf{house}  \\
\reenttype{repetition} &   a repeated name or non-pronominal phrase (non-anaphoric coreference) & The U.S. promotes \uline{American} \textbf{goods} \\
\hline
\reenttype{coordination} &   coordination of two or more phrases sharing an argument & They cheered \uline{and} \textbf{celebrated} \\
\hline
\reenttype{control} &    control verbs, control nouns, or control adjectives & I was \uline{afraid} to \textbf{speak up}  \\
\reenttype{adjunct control} &  control within an adjunct phrase & I left \uline{to} \textbf{buy} some milk; Mary cooked \uline{while} \textbf{listening} to music \\
\reenttype{unmarked adjunct control} &  control within an adjunct phrase with only a bare verb and no subordinating conjunction & Mary did her homework \uline{\textbf{listening}} to music \\
\reenttype{comparative control} & a comparative construction &  Be \uline{as} \textbf{objective} as possible \\
\hline
\reenttype{pragmatic} &  Reentrancies that must be resolved using context & John met up with a \textbf{\uline{friend}} \\
\end{tabular}
\caption{Reentrancy types with examples. For each reentrant node, one of its incoming edges is labeled \reenttype{primary} and the others are labeled with one of the above reentrancy types. In the examples, the word aligned to an edge labeled with the specified type is underlined, and the word aligned to the parent of that edge is bolded.}
\label{tab:reenttypes}
\end{table*}

The reentrancy layer ``explains'' the cause of each reentrancy as follows: for the incoming edges of a reentrant node, one of these edges is designated as \reenttype{primary}---this is usually the first mention of the entity in a local surface syntactic attachment, e.g.~the argument of a control predicate like \w{want} doubles as an argument of an embedded clause predicate.\hide{\footnote{If a pronoun has a non-pronominal antecedent in the sentence, the antecedent is preferred as \reenttype{primary}.}}
The remaining incoming edges to a reentrant node are aligned to a \textbf{reentrancy trigger} and labeled with one of 8~\textbf{reentrancy types}: \textit{coref, repetition, coordination, control, adjunct control, unmarked adjunct control, comparative control,} and \textit{pragmatic}. These are illustrated in \cref{tab:reenttypes}. 
These types, adapted from \citeposs{szubert-20} classification,\hide{\footnote{We add several categories for rare reentrancy types (unmarked adjunct control, and comparative control), conflate verbal, nominal, and adjectival control into one category, and draw a distinction between two types of coreference (coref and repetition). Our categories generally correspond to tests used to identify reentrancy types. We think our categories have better coverage.}} correspond to different linguistic phenomena leading to AMR reentrancies---anaphoric and non-anaphoric coreference, coordination, control, etc.
The trigger is the word that most directly signals the reentrancy phenomenon in question.
For the example in \cref{fig:alignment_layers}, the control verb \w{want} is aligned to the embedded predicate--argument relation and typed as \reenttype{control}, while the pronoun \w{they} serves as the trigger for the third instance of \amr{p} in \pex{when \uline{they} graduate}.
%Further examples appear in \cref{tab:reenttypes}. %, where each reentrancy's trigger is underlined.

\subsection{Validation}

To validate the annotation scheme we elicited two gold-standard annotations for 40 of the test sentences described in \cref{sec:data-release} and measured interannotator agreement.\footnote{Both annotators are Ph.D.\ students with backgrounds in linguistics. One annotator aligned all development and test sentences; the other aligned a subset of 40 test sentences.}
Interannotator exact-match F1 scores were 94.54 for subgraphs, 90.73 for relations, 76.92 for reentrancies, and 66.67 for duplicate subgraphs (details in \cref{sec:IAA}).

% \subsection{Annotation}

% In addition to our automatic alignments, we release 350 sentences (150 dev and 200 test) of gold, hand-aligned data for evaluation, as well as our annotation guidelines. The 200 test sentences were annotated from scratch.\footnote{Both annotators are Ph.D.\ students with backgrounds in linguistics. One annotator aligned 200 sentences; the other aligned a subset of 40.} We stress that no preprocessing apart from tokenization is required to prepare the test sentences and AMRs for human annotation. The 150 development sentences were first automatically aligned and then hand-corrected. 

%\topic{Interannotator Agreement} We calculated interannotator agreement on a subset of 40 test sentences. Interannotator exact-match F1 scores were 94.54 for subgraphs, 90.73 for relations, 76.92 for reentrancies, and 66.67 for duplicate subgraphs (details in \cref{sec:IAA}).

\section{Released Data}\label{sec:data-release}

We release a dataset\footnote{\url{https://github.com/ablodge/leamr}} of the four alignment layers reflecting correpondences between English text and various linguistic phenomena in gold AMR graphs---subgraphs, relations (including argument structures), reentrancies (including coreference, control, etc.), and duplicate subgraphs.

\textbf{Automatic alignments} cover
the $\approx$60,000 sentences of the LDC2020T02 dataset \citep{amr2020} and $\approx$1,500 sentences of \textit{The Little Prince}.

We manually created \textbf{gold alignments} for evaluating our automatic aligner, split into a development set (150~sentences) and a test set (200~sentences).\footnote{Our test set consists of sentences from the test set of \citet{szubert2018} but with AMRs updated to the latest release version. This test set contains a mix of English sentences drawn from the LDC data and \textit{The Little Prince}---some sampled randomly, others hand-selected---as well as several sentences constructed to illustrate particular phenomena.} The test sentences were annotated from scratch; the development sentences were first automatically aligned and then hand-corrected.
We stress that no preprocessing apart from tokenization is required to prepare the test sentences and AMRs for human annotation.
We also release our annotation guidelines as a part of our data release.

\section{LEAMR Aligner}\label{sec:aligner}

We formulate statistical models for the alignment layers described above---\textbf{subgraphs}, \textbf{duplicate subgraphs}, \textbf{relations}, and \textbf{reentrancies}---and use the Expectation-Maximization (EM) algorithm to estimate probability distributions without supervision, with a decoding procedure that constrains aligned units to obey structural requirements.
%align components of an AMR to English spans using a probabilistic score while constraining the spans that can be aligned to a set of legal candidates in order to preserve the AMR structure. This allows our aligner to produce statistically learned alignments to connected substructures of the AMR. 
In line with \citet{jamr,jamr2}, we use rule-based preprocessing to align some substructures using string-matching, morphological features, etc.

\begin{figure}
    \centering\small
    \begin{verbatim}
(h / have-degree-91 
    :ARG1 (h2 / house :location (l / left))
    :ARG2 (b / big) 
    :ARG3 (m / more) 
    :ARG4 (h3 / house :location (r / right)))\end{verbatim}
    \caption{AMR for the sentence ``The house$_1$ on the left is bigger than the house$_2$ on the right.''}
    \label{fig:distexample}
\end{figure}

Before delving into the models and algorithm, we motivate two important characteristics:

\topic{Structure-Preserving} Constraints on legal candidates during alignment ensure that at any point only connected substructures may be aligned to a span. Thus, while our aligner is probabilistic like the ISI aligner, it has the advantage of preserving the AMR graph structure. 

\topic{Projection Distance} The scores calculated for an alignment take into account a distance metric designed to encourage locality---tokens that are close together in a sentence are aligned to subtructures that are close together in the AMR graph. We define the \textit{projection distance} \(\textit{dist}(\amr{n1},\amr{n2})\) between two neighboring nodes \amr{n1} and \amr{n2} to be the signed distance in the corresponding sentence between the span aligned to \amr{n1} and the span aligned to \amr{n2}. This motivates the model to prefer alignments whose spans are close together when aligning nodes which are close together---particularly useful when a word occurs twice with identical subgraphs. Thus, our aligner relies on more information from the AMR graph structure than other aligners (note that the ISI system linearizes the graph). %\citep[contrast][who align tokens to a linearization of the AMR graph]{pourdamghani2014}. 
%Given an alignment function $f$, we model the projection distance between two nodes \amr{n1} and \amr{n2} as \(dist(\amr{n1},\amr{n2}) = f(\amr{n1}) - f(\amr{n2})\) where $f$ maps each node to the index of the span aligned to that node. 
%For $\textit{dist}$ we use the signed distance to preserve ordering information.
%where negative (or \textit{positive}) distance indicates the aligned span of \amr{n1} is on the left (or \textit{right}) compared to \amr{n2}.
%
%Consider the sentence ``The house$_1$ on the left is bigger than the house$_2$ on the right.'', where there are two \w{house} tokens and the AMR will contain two \amr{house} nodes. For each possible alignment, the translation probability \(P_{\text{align}}(\amr{house}\mid\w{house})\) will be the same, thus resulting in an ambiguity. To address this, we use an additional probability to model projection distance \(P_{\text{dist}}(dist(\amr{n1},\amr{n2}))\). %which considers the distance between tokens aligned to each neighboring node. 
%Assuming there are alignments \w{left} $\rightarrow$ \amr{l/left} and \w{right} $\rightarrow$ \amr{r/right}, the projection distance will be minimized if \w{house$_1$} is aligned to the \amr{house} node which is a neighbor of \amr{l/left} since the tokens \w{house$_1$} and \w{left} are close together. 
Further details are given in \cref{sec:subgaphaligner}.

\subsection{Overview}\label{sec:algorithm}

% \begin{algorithm}\small
% \begin{algorithmic}[1]
%     \Function{AlignSentence}{$\mathbf{w}$, $\mathcal{G}$}
%         \State{$\mathbf{s} \gets $ \textsc{SegmentSpans}($\mathbf{w}$)}
%         \State{$\mathcal{A}_{\text{subgraph}} \gets $ \textsc{AlignSubgraphs}($\mathbf{s}$, $\mathcal{G}$)} \Comment{\Cref{pseudocode}}
%         \State{$\mathcal{A}_{\text{rel}} \gets $ \textsc{AlignRelations}($\mathbf{s}$, $\mathcal{G}$, $\mathcal{A}_{\text{subgraph}}$)}
%         \State{$\mathcal{A}_{\text{reent}} \gets $ \textsc{AlignReent}($\mathbf{s}$, $\mathcal{G}$, $\mathcal{A}_{\text{subgraph}}$, $\mathcal{A}_{\text{rel}}$)}
%         \State{\Return{$\mathbf{s}$, $\mathcal{A}$}}
%     \EndFunction
% \end{algorithmic}
% \caption{Alignment prediction pipeline. 
% \textsc{AlignRelations} and \textsc{AlignReent} use essentially the same search procedure as \textsc{AlignSubgraphs} but with different candidates (respectively: unaligned edges; reentrant edges), different criteria for legal candidates, and different scoring functions.}
% \end{algorithm}

%We use this basic algorithm to build four alignment layers---\textbf{subgraphs}, \textbf{duplicate subgraphs}, \textbf{relations}, and \textbf{reentrancies}---described in greater detail below.

% We define a generative probabilistic model to score alignments between token spans and AMR subgraphs, subject to structural constraints enforced in a greedy decoding procedure.
% Parameters are learned via EM with initialization based on rules.
% Hyperparameters are tuned on development data.

\Cref{pseudocode} illustrates our base algorithm in pseudocode. The likelihood for a sentence can be expressed as a sum of per-span alignment scores:
we write the score of a full set of a sentence's subgraph alignments $\mathcal{A}$ as
\begin{equation}\label{eq:mainscore}
\textit{Score}(\mathcal{A} \mid \mathcal{G}, \mathbf{w}) = \prod_{i=1}^N{\textit{score}(\langle \mathbf{g}_i,s_i \rangle \mid \mathcal{G}, \mathbf{w})}
\end{equation}
%\[ \textit{Score}(\mathcal{A} \mid \mathcal{G}, \mathbf{w}) = \sum_{i=1}^N{\textit{score}(\langle \mathbf{g}_i,s_i \rangle, \mathcal{N}(\mathbf{g}_i; \mathcal{G}, \mathcal{A}) \mid \mathcal{G}, \mathbf{w})} \]
%The decoding objective for a sentence factorizes by aligned (subgraph, span) pairs:
%\[ \argmax_{\mathbf{g},\mathbf{s}} \textit{Score}(\mathbf{g}, \mathbf{s}) = \argmax_{\mathbf{g},\mathbf{s}} \sum_{i=1}^N{\textit{score}(g_i,s_i)} \]
where $\mathbf{s}$ are $N$ aligned spans in the sentence $\mathbf{w}$, and $\mathbf{g}$ are sets of subgraphs of the AMR graph $\mathcal{G}$ aligned to each span.
%$\mathcal{N}(\mathcal{G}, \mathcal{A})$ denotes pairs of neighboring nodes in $\mathcal{G}$ that belong to differently aligned subgraphs in $\mathcal{A}$; $s_{g_n}$ is the span aligned to the subgraph containing node $n$, and $d(\cdot,\cdot)$ is the distance between two spans. 
For relations model and the reentrancies model, each $\textbf{g}_i$ consists of relations rather than subgraphs.
Henceforth we assume all alignment scores are conditioned on the sentence and graph and omit $\mathbf{w}$ and $\mathcal{G}$ for brevity. The $score(\cdot)$ component of \cref{eq:mainscore} is calculated differently for each of the three models detailed below.

% \begin{multline}
% \textit{Score}(\mathcal{A} \mid \mathcal{G}, \mathbf{w}) = \prod_{i=1}^N{\textit{Score}_\text{align}(\langle \mathbf{g}_i,s_i \rangle \mid \mathcal{G}, \mathbf{w})}
% \\
% \cdot \prod_{n,n'\in \mathcal{N}(\mathcal{G},\mathcal{A})} P_{\text{dist}}(d(s_{g_n},s_{g_{n'}})) 
% \end{multline}
% %\[ \textit{Score}(\mathcal{A} \mid \mathcal{G}, \mathbf{w}) = \sum_{i=1}^N{\textit{score}(\langle \mathbf{g}_i,s_i \rangle, \mathcal{N}(\mathbf{g}_i; \mathcal{G}, \mathcal{A}) \mid \mathcal{G}, \mathbf{w})} \]
% %The decoding objective for a sentence factorizes by aligned (subgraph, span) pairs:
% %\[ \argmax_{\mathbf{g},\mathbf{s}} \textit{Score}(\mathbf{g}, \mathbf{s}) = \argmax_{\mathbf{g},\mathbf{s}} \sum_{i=1}^N{\textit{score}(g_i,s_i)} \]
% where $\mathbf{s}$ are $N$ aligned spans in the sentence $\mathbf{w}$, and $\mathbf{g}$ are sets of subgraphs of the AMR graph $\mathcal{G}$ aligned to each span.
% $\mathcal{N}(\mathcal{G}, \mathcal{A})$ denotes pairs of neighboring nodes in $\mathcal{G}$ that belong to differently aligned subgraphs in $\mathcal{A}$; $s_{g_n}$ is the span aligned to the subgraph containing node $n$, and $d(\cdot,\cdot)$ is the distance between two spans. Henceforth we assume all alignment scores are conditioned on the sentence and graph and omit $\mathbf{w}$ and $\mathcal{G}$ for brevity.

% The score for a single alignment would then be:
% \begin{equation}
%     score(\langle g, s\rangle) = P_{\text{align}}(g\mid s; \theta_1)  \cdot \textit{R}(g, s)
% \end{equation}
% with an optional regularizer $R$---see inductive bias under \cref{sec:subgaphaligner}.

\begin{algorithm*}\small
\begin{algorithmic}[1]
    \Function{AlignSubgraphs}{spans, amr}
        \State{alignments $\gets$ dict()} \Comment{map from span to an ordered list of aligned subgraphs}
        \State{unaligned\_nodes $\gets$ get\_unaligned\_nodes(amr, alignments)}
        \While{$|\text{unaligned\_nodes}| > 0$}
            \State $\Delta$scores $\gets$ []
            \State candidate\_s\_g\_pairs $\gets$ []
            \For{n $\in$ unaligned\_nodes}
            	\State candidate\_spans $\gets$ get\_legal\_alignments(n, alignments) 
            	\For{span, i\_subgraph $\in$ candidate\_spans} \Comment{either there is an edge between \emph{n} and the indicated subgraph already aligned to \emph{span}, or \emph{i\_subgraph} would be a new subgraph consisting of \emph{n}}
            		\State current\_aligned\_nodes $\gets$ alignments[span][i\_subgraph] \Comment{$\emptyset$ if this would be a new subgraph}
            		\State new\_aligned\_nodes $\gets$ current\_aligned\_nodes $\cup$ $\{\text{n}\}$
            		\State $\Delta$score $\gets$ get\_score(span, new\_aligned\_nodes, alignments)
            		\State ~~~~~~~~~~~~$-$ get\_score(span, current\_aligned\_nodes, alignments)
            		\Comment{change from adding \emph{n} into a subgraph aligned to \emph{span}; \emph{get\_score} queries \emph{$\textit{score}(\langle g, s\rangle)$} and multiplies \emph{$\lambda_{\text{dup}}$} \textit{if} \emph{$\text{i\_subgraph}>1$}}
            		%\If{i\_subgraph $>$ 1}
            		%    \State score $\gets$ $\lambda_{\text{dup}} \cdot \text{score}$ \Comment \textit{penalize duplicate subgraph alignments}
            		%\EndIf
            		\State $\Delta$scores.add($\Delta$score)
            		\State candidate\_s\_g\_pairs.add((span, new\_aligned\_nodes, i\_subgraph))
            	\EndFor
            \EndFor
            %\State best $\gets$ argmax(scores)
            %\State best\_span $\gets$ candidate\_spans[argmax(scores)] % won't work because candidate\_spans is reassigned for each n
            \State span$^*$, subgraph$^*$, i\_subgraph$^*$ $\gets$ candidate\_s\_g\_pairs[argmax($\Delta$scores)] \Comment{\textit{update having the best impact on score (equivalently, maximizing sum of scores across individual aligned spans)}}
            %\State align(alignments, best\_n, best\_span)
            \State alignments[span$^*$][i\_subgraph$^*$] $\gets$ subgraph$^*$
            \State{unaligned\_nodes $\gets$ get\_unaligned\_nodes(amr, alignments)}
        \EndWhile
        \State{\Return{alignments}}
    \EndFunction
    % functions called: get_unaligned_nodes, get_legal_alignments, get_score
\end{algorithmic}
\caption{Procedure for greedily aligning all nodes to spans using a scoring function that decomposes over (span, subgraph) pairs. (Scores are expressed in real space but the implementation is in log space.)}
\label{pseudocode}
\end{algorithm*}

\topic{Alignment Pipeline}
Alignment proceeds in the following phases, with each phase depending on the output of the previous phase: 
\begin{enumerate}
    \item \textit{Preprocessing:} Using external tools we extract lemmas, %\footnote{Lemmas are used in all phases to normalize token spans.} 
    parts of speech, and coreference. % for each sentence. %Part of speech and coreference are used in some layers to help identify legal candidates.
    \item \textit{Span Segmentation:} Tokens are grouped into spans using a rule-based procedure (\cref{sec:appendix-spans}).
    \item \textit{Align Subgraphs \& Duplicate Subgraphs:} We greedily identify subgraph and duplicate subgraph alignments in the same alignment phase (\cref{sec:subgaphaligner}). 
    \item \textit{Align Relations:} Relations not belonging to a subgraph are greedily aligned in this phase, using POS criteria to identify legal candidates (\cref{sec:rel}).
    \item \textit{Align Reentrancies:} Reentrancies are aligned in this phase, using POS and coreference in criteria for identifying legal candidates (\cref{sec:reent}).
\end{enumerate}

The three main alignment phases use different models with different parameters; they also have their own preprocessing rules used to identify some alignments heuristically (\cref{sec:subgraph-preproc,sec:rel-preproc,sec:reent-preproc}).\footnote{79\% of nodes and 89\% of edges are aligned by rules. We believe this is why in practice, EM performs well without random restarts.}
In training, parameters for each phase are iteratively learned and used to align the entire training set by running EM to convergence before moving on to the next phase.
At test time, the pipeline can be run sentence-by-sentence.

\topic{Decoding}
%Pseudocode for our substructure-aware alignment algorithm appears as \cref{pseudocode}. 
The three main alignment phases all use essentially the same greedy, substructure-aware search procedure.
This searches over node--span candidate pairs based on the scoring function modeling the compatibility between a subgraph (or relation) $g$ and span $s$, which we denote $\textit{score}(\langle g, s\rangle)$.
For each unaligned node (or edge), we identify a set of legal candidate alignments using phase-specific criteria.
The incremental score improvement of adding each candidate---either extending a subgraph/set of relations already aligned to the span, or adding a completely new alignment---is calculated as
%Given a scoring function for single alignments, we use a score differential to choose the best next alignment, making decisions greedily. The score differential is the difference in scores between a candidate alignment and the current alignment with the same span. 
as $\Delta \textit{score} = \textit{score}(\langle g_0\cup \{n\}, s\rangle) - \textit{score}(\langle g_0, s\rangle)$,
where $g_0$ is the current aligned subgraph, $s$ is the span, and $n$ is an AMR component being considered. %Each candidate alignment receives an alignment score. 
Of the candidates for all unaligned nodes, the node--span pair giving the best score improvement is then greedily selected to add to the alignment. 
%(Once a node is aligned to a span it will never be removed from that alignment.)
This is repeated until all nodes have been aligned (even if the last ones decrease the score).
The procedure is detailed in \cref{pseudocode} for subgraphs; the relations phase and the reentrancies phase use different candidates (respectively: unaligned edges; reentrant edges), different criteria for legal candidates, and different scoring functions.

\subsection{Aligning Subgraphs}\label{sec:subgaphaligner}

The score assigned to an alignment between a span and subgraph is calculated as $score(\langle g, s\rangle) =$
\begin{equation}
P_{\text{align}}(g\mid s; \theta_1) \cdot \prod_{d_i \in D} P_{\text{dist}}(d_i; \theta_2)^{\frac{1}{\abs{D}}} \cdot \textit{IB}(g, s)
\end{equation}
%\ajb{This is not a probability because of $IB$ which may is based on PMI}\nss{oh, so it's not a fully probabilistic model?}
where $g$ is a subgraph, $s$ is a span, $d_i$ is the projection distance of $g$ with its $i$th neighboring node, and $\theta_1$ and $\theta_2$ are model parameters which are updated after each iteration. The subgraph $g$ is 
%given a label\footnote{with inverse edges normalized} which describes all concepts and edges contained in the subgraph. 
represented in the model as a bag of concept labels and (parent concept, relation, child concept) triples. %, one per edge.

The distributions \(P_{\text{align}}\)  and \(P_{\text{dist}}\) are inspired by IBM Model~2 \citep{brown-88}, and can be thought of as graph-theoretic extensions of translation ($\text{align}$) and alignment ($\text{dist}$) probabilities.
$\textit{IB}$ stands for \emph{inductive bias}, explained below.

\topic{Legal Candidates} For each unaligned node $n$, the model calculates a score for spans of three possible categories: 
1)~unaligned spans; 
2)~spans aligned to a neighboring node (in this case, the aligner considers adding $n$ to an existing subgraph if the resulting subgraph would be connected);
3)~spans aligned to a node with the same concept as $n$ (this allows the aligner to identify duplicate subgraphs---candidates in this category receive a score penalty because duplicates are quite rare, so they are generally the option of last resort).

Limiting the candidate spans in this way ensures only connected, plausible substructures of the AMR are aligned. To form a multinode subgraph alignment \w{t$_1$} $\rightarrow$ \amr{n1 :rel n2}, the aligner could first align \amr{n1} to an unaligned span \w{t$_1$}, then add \amr{n2}, which is a legal candidate because \w{t$_1$} is aligned to a neighboring node of \amr{n2} (ensuring a connected subgraph). 

\topic{Distance} 
We model the probability of the projection distance \(P_{\text{dist}}(d; \theta_2)\) using a Skellam distribution, which is the difference of two Poisson distributed random variables $D = N_1 - N_2$ and can be positive or negative valued. Parameters are updated based on alignments in the previous iteration. \hide{\nss{When scoring a candidate alignment,}\ajb{remove this?}} For each aligned neighbor $n_i$ of a subgraph $g$, we calculate \(P_{\text{dist}}(dist(g, n_i); \theta_2)\) and take the geometric mean of probabilities as \(P_{\text{dist}}\).\hide{\footnote{In the global objective, we have written the distance as an average over all pairs of neighboring nodes in adjacent subgraphs.
In practice, we approximate this by computing, when scoring the incremental addition of a new subgraph alignment, the average span distance between nodes in that alignment and neighboring nodes in previously aligned subgraphs.}}

\topic{Null alignment} The aligner models the possibility of a span being unaligned using a fixed heuristic: 
\begin{equation}
P_{\text{align}}(\emptyset\mid s) = \max\{\textit{rank}(s)^{-\frac{1}{2}},0.01\}
\end{equation}
where $\textit{rank}$ assigns $1$ to the most frequent word, $2$ to the 2nd most frequent, etc. Thus, the model expects that very common words are more likely to be null-aligned and rare words should almost always be aligned.\footnote{We allow several exceptions. For punctuation, words in parentheses, and spans that are coreferent to another span, the probability is $0.5$. For repeated spans, the probability is $0.1$.} 

\topic{Factorized Backoff} So that the aligner generalizes to unseen subgraph--span pairs, where $P_{\text{align}}(g \mid s)=0$, we use a backoff factorization into components of the subgraph. %The factorized model utilizes empirical probabilities of component nodes and edges cooccuring with the span in training sentences: 
In particular, the factors are empirical probabilities of (i)~an AMR concept given a span string in the sentence, and (ii)~a relation and child node concept given the parent node concept and span string.
These cooccurrence probabilities $\hat{p}$ are estimated directly from the training sentence/AMR pairs (irrespective of latent alignments).
The product is scaled by a factor $\lambda$.
%$P_{\text{factorized}}$ of a subgraph and span pair is calculated as a product of probabilities for each node or edge conditioned on the span, with a penalty rate $\lambda$. 
E.g., for a subgraph \amr{n1 :rel1 n2 :rel2 n3}, % and span $s$, 
where $\amr{c}_{\amr{n}}$ is the concept of node \amr{n}, we have
\begin{multline}
\hspace{-.8em} P_{\text{factorized}}(g\mid s) =
\lambda \cdot \hat{p}(\amr{c}_{\amr{n1}}\mid s) 
\cdot \hat{p}(\amr{:rel1}, \amr{c}_{\amr{n2}}\mid \amr{c}_{\amr{n1}}, s)  \\
\cdot \hat{p}(\amr{:rel2}, \amr{c}_{\amr{n3}}\mid \amr{c}_{\amr{n1}}, s)
\end{multline}
\hide{\nss{I assume these probabilities actually look at the concepts? $\textit{C}_{\amr{n1}}$, etc.}}

\topic{Inductive bias}
Lastly, to encourage good initialization, the score function includes an inductive bias which does not depend on EM-trained parameters. This inductive bias is based on the empirical probability of a node occurring in the same AMR with a span in the training data. We calculate inductive bias as an average of exponentiated PMIs \(\frac{1}{N}\sum_i\exp(\textit{PMI}(n_i, s))\), where $N$ is the number of nodes in $g$, $n_i$ is the $i$th node contained in the subgraph, and $\textit{PMI}$ is the PMI of $n_i$ and $s$.

\topic{Aligning Duplicate Subgraphs} On rare occasion a span should be aligned to multiple subgraphs (\cref{sec:duplicates-layer}). 
To encourage the model to align a different span where possible, there is a constant penalty $\lambda_{\text{dup}}$ for each additional subgraph aligned to a span beyond the first.
Thus the score for a span and its subgraphs is computed as:
\begin{equation}
\textit{score}(\langle \mathbf{g}, s\rangle) = \lambda_{\text{dup}}^{|\mathbf{g}|-1} \prod_{g \in \mathbf{g}} \textit{score}(\langle g, s\rangle)
\end{equation}
% \ajb{
% The idea is that there is a numerical penalty for aligning something as a duplicate. It looks like lambda+S(duplicate\_align)-S(current\_align) where lambda is a penalty. The aligner will only treat something as a duplicate of a span if aligning to that span is the least bad possible alignment. } 

\begin{table*}[!ht]
    \centering\small
    \begin{tabular}{|c|ccc|ccc|c|c|c|c|}
    \cline{2-9}
        \multicolumn{1}{c|}{} & \multicolumn{3}{c|}{\textbf{Exact Align}} & \multicolumn{3}{c|}{\textbf{Partial Align}} & \textbf{Spans} & \textbf{Coverage}\\
    %\cline{2-9}
        \multicolumn{1}{c|}{} & P & R & F1 & P & R & F1 & F1 &\\
    \cline{2-9}
    \multicolumn{1}{c}{} & \multicolumn{8}{c}{Subgraph Alignments ($N=1707$)}\\
    \hline
        Our system & 93.91 & 94.02 & 93.97 & 95.69 & 95.81 & 95.75 & 96.05 & 100.0 \\	
        JAMR & 87.21 & 83.06 & 85.09 & 90.29 & 85.99 & 88.09 & 92.38 & \hphantom{0}91.1 \\
        ISI & 71.56 & 68.24 & 69.86 & 78.03 & 74.54 & 76.24 & 86.59 & \hphantom{0}78.7 \\
        TAMR (91 sentences) & 85.68 & 83.38 & 84.51 & 88.62 & 86.24 & 87.41 & 93.64 & \hphantom{0}94.9 \\ 
    \hline
    \multicolumn{1}{c}{} & \multicolumn{8}{c}{Relation Alignments ($N=1263$)}\\
    \hline
    Our system & 85.67 & 85.37 & 85.52 & 88.74 & 88.44 & 88.59 & 95.41 & 100.0 \\
    ISI & 59.28 & \hphantom{0}8.51 & 14.89 & 66.32 & \hphantom{0}9.52 & 16.65 & 83.09 & \hphantom{00}9.8 \\
  \hline
  \multicolumn{1}{c}{} & \multicolumn{8}{c}{Reentrancy Alignments ($N=293$)}\\
  \hline
    Ours (labeled) & 55.75 & 54.61 & 55.17 & - & - & - & - & 100.0 \\
    Ours (unlabeled) & 62.72 & 61.43 & 62.07 & - & - & - & - & 100.0 \\
    \hline
    
\multicolumn{1}{c}{} & \multicolumn{8}{c}{Duplicate Subgraph Alignments ($N=17$)}\\
  \hline
    Our system & 66.67 & 58.82 & 62.50 &  70.00 & 61.76 & 65.62 & - & 100.0 \\
    \hline
    \end{tabular}
    \caption{Main results on the test set.
    $N$ represents the denominator of exact alignment recall.
    There are 2860 gold spans in total, 41\% of which are null-aligned and 0.6\% of which are aligned to multiple subgraphs. 95\% of the spans consist of a single token, and 49\% of spans are aligned to a single subgraph consisting of a single node.}
    \label{tab:mainresults}
\end{table*}

\subsection{Aligning Relations}\label{sec:rel}

For a given relation alignment between a span and a collection of edges, we calculate a score as follows:
\begin{multline}\label{eq:rel}
\hspace{-.7em} \textit{score}(\langle a, s\rangle)  = P_{\text{align}}(a\mid s;\theta_3) \cdot \prod_{d_i \in D_1} P_{\text{dist}}(d_i ; \theta_4)^{\frac{1}{\abs{D_1}}} \\
\cdot \prod_{d_j \in D_2} P_{\text{dist}}(d_j ; \theta_5)^{\frac{1}{\abs{D_2}}}
\end{multline}
where $a$ is the argument structure (the collection of aligned edges), $s$ is a span, $D_1$ is the projection distances of each edge and its parent, and $D_2$ is the projection distances of each edge and its child. %, and $\theta_3$, $\theta_4$, and $\theta_5$ are model parameters which are updated after each iteration. 
The collection of edges $a$ is given a normalized label which represents the relations contained in the alignment (distinguishing incoming versus outgoing relations, and normalizing inverse edges).

\topic{Legal Candidates} There are two kinds of candidate spans for relation alignment. First, previously unaligned spans\footnote{We constrain these to particular parts of speech: prepositions (IN), infinitival to (TO), possessives (POS), and possessive pronouns (PRP\$)\hide{\nss{also include WP\$ for ``whose''?}}. Additionally, only spans that are between the spans aligned to the parent and any descendent of child nodes of the relation (and are not between the child's aligned span and any of its descendants' spans) are allowed. This works well in practice for English.} (with no relation or subgraph alignments), e.g.~prepositions and subordinating conjunctions such as \w{in} $\rightarrow$ \amr{:location} or \w{when} $\rightarrow$ \amr{:time}. Second, any spans aligned to the relation's parent or child in the subgraph layer: this facilitates alignment of argument structures such as \w{give} $\rightarrow$ \amr{:ARG0 :ARG1 :ARG2}. Additionally, we constrain certain types of edges to only align with the parent and others to only align with the child. 

\topic{Distance} For relations there are potentially two distances of interest---the projected distance of the relation from its parent and the projected distance of the relation from its child. We model these separately as \textit{parent distance} and \textit{child distance} with distinct parameters. To see why this is useful, consider the sentence ``Should we meet at the restaurant or at the office?'', where each \w{at} token should be aligned to a \amr{:location} edge. In English, prepositions like \w{at} precede an object and follow a governor. Thus parent distance tends to be to the left (negative valued) while child distance tends to be to the right (positive valued).

% Lastly, the score function includes an inductive bias which is not a function of the parameters to encourage good initialization. This inductive bias is based on the empirical probability of a node occurring in the same AMR with a span in the training data. We calculate inductive bias as follows:
% \[\textit{IB}(g, s) = \frac{1}{N}\sum_i\exp(\textit{PMI}(n_i, s))\]
% where $g$ is a subgraph, $s$ is a span, $N$ is the number of nodes in $g$, $n_i$ is the $i$th node contained in the subgraph, and $PMI$ is the point-wise mutual information of $n_i$ and $s$.

\subsection{Aligning Reentrancies}\label{sec:reent}

The probability of a reentrancy alignment is similar to \cref{eq:rel}, but with an extra variable 
%$type$ representing the type of reentrancy assigned to the alignment: 
for the reentrancy type:
$\textit{score}(\langle r, s, \textit{type}\rangle) = $
\begin{equation}
P_{\text{align}}(r,\textit{type}\mid s;\theta_6) \cdot P_{\text{dist}}(d_1 ; \theta_7) \cdot P_{\text{dist}}(d_2 ; \theta_8)
\end{equation}
where $r$ is the role label of the reentrant edge. %, $s$ is a span, $d_1$ is the projection distance of the edge and its parent, $d_2$ is the projection distance of the edge and its child, and $\theta_6$, $\theta_7$, and $\theta_8$ are model parameters which are updated after each iteration. 

\topic{Legal Candidates} There are 8 reentrancy types (\cref{sec:reentrancies-layer}). For each type, a rule-based test determines if a span and edge are permitted to be aligned. The 8 tests use part of speech, the structure of the AMR, and subgraph and relation alignments.
%to determine if an alignment is permitted. 
A span may be aligned (rarely) to multiple reentrancies, but these alignments are scored separately.

\section{Experimental Setup}

Sentences are preprocessed with the Stanza library \citep{qi2020stanza} to obtain lemmas, part-of-speech tags, and named entities. We identify token spans using a combination of named entities and a fixed list of multiword expressions (details are given in \cref{sec:appendix-spans}). Coreference information, which is used to identify legal candidates in the reentrancy alignment phase, is obtained using NeuralCoref.\footnote{\url{https://github.com/huggingface/neuralcoref}} Lemmas are used in each alignment phase to normalize representation of spans, while parts of speech and coreference are used to restrict legal candidates in the relation and reentrancy alignment phases. We tune hyperparameters, including penalties for duplicate alignments and our factorized backoff probability, on the development set.

\section{Results}\label{sec:results}
\Cref{tab:mainresults} describes our main results on the 200-sentence test set (\cref{sec:data-release}),
reporting exact-match and partial-match alignment scores as well as span identification F1 and coverage\footnote{A previous draft of this work reported lower scores on relations before a constraint was added to improve the legal candidates for relation alignment.}.
The partial alignment evaluation metric is designed to be more forgiving of arbitrary or slight differences between alignment systems. We argue that this metric is more comparable across alignment systems. It assigns partial credit equal to the product of Jaccard indices \(\frac{\abs{N_1\cap N_2}}{\abs{N_1 \cup N_2}}\cdot\frac{\abs{T_1 \cap T_2}}{\abs{T_1 \cup T_2}}\) for nodes (or edges) and tokens respectively. This partial credit is calculated for each gold alignment and the closest matching predicted alignment with nodes (or edges) $N_1$ and $N_2$ and tokens $T_1$ and $T_2$. Coverage is the percentage of relevant AMR components that are aligned. 

Our aligner shows improvements over previous aligners in terms of coverage and accuracy even when using a partial credit metric for evaluation. We demonstrate greater coverage, including coverage of phenomena not aligned by previous systems.

\Cref{tab:relationresults} shows detailed results for relation subtypes and reentrancy subtypes. Here, we see room for improvement. In particular, ISI outperforms our system at aligning single relations. Our reentrancy aligner lacks a baseline to compare to, but the breakdown of results by type suggest there are several categories of reentrancies where scores could be improved.

\begin{table}[]
    \centering\small
\resizebox{1.01\columnwidth}{!}{
    \begin{tabular}{@{}|l|ccc|@{}H@{}H@{}H@{}}
    \cline{2-4}
         \multicolumn{1}{@{}c}{} & \multicolumn{3}{|c|}{\textbf{Exact Align}} \\ %& \multicolumn{3}{H|}{\textbf{Partial Align}}\\
         \multicolumn{1}{@{}c}{} & \multicolumn{1}{|c}{P} & R & F1 & P & R & F1\\
         \cline{1-4}
         \multicolumn{4}{@{}|c|}{\textbf{Relation Alignments Breakdown}}\\
         \cline{1-4}
    % \cline{2-4}
    %\multicolumn{1}{c}{} & \multicolumn{6}{c}{Relation Alignments ($N=1345$)}\\
    % \cline{1-4}
    Our system: all (1163)  & 85.67 & 85.37 & 85.52 & 88.74 & 88.44 & 88.59 \\
    \dots single relations (121) & 53.49 & 56.56 & 54.98 & 54.91 & 58.06 & 56.44 \\
    \dots argument structures (1042) & 89.67 & 88.73 & 89.20 & 92.82 & 91.85 & 92.33 \\[5pt]
    ISI: all (1163) & 59.28 & \hphantom{0}8.51 & 14.89 & 66.32 & \hphantom{0}9.52 & 16.65 \\
     \dots single relations (121) & 82.89 & 52.07 & 63.96 & 83.22 & 52.27 & 64.21 \\
    \dots argument structures (1042) & 39.56 & \hphantom{0}3.45 & \hphantom{0}6.35  & 52.20 & \hphantom{0}4.56 & \hphantom{0}8.38 \\
%   \cline{1-4}
%   \cline{2-4}
\cline{1-4}
        \multicolumn{4}{@{}|c|}{\textbf{Reentrancy Alignments Breakdown}}\\
%    \multicolumn{1}{c}{} & \multicolumn{3}{c}{Reentrancy Alignments ($N=293$)}\\
\cline{1-4}
    Our system: all (293) & 62.37 & 61.09  & 61.72 \\
    \dots primary (128) & 79.37 & 78.12  & 78.74 \\
    \dots coref (41)& 57.14 & 58.54  & 57.83 \\
    \dots control (36)& 73.08 & 52.78 & 61.29 \\
    \dots coordination (29)& 57.14 & 58.54  & 57.83 \\
    \dots pragmatic (25) & 20.93 & 36.00 & 26.47 \\
    \dots adjunct control (15)& 100.00 & \hphantom{0}6.67  & 12.50 \\
    \dots repetition (13) & 60.00 & 46.15 & 52.17  \\
    \dots comparative control (5) &  0.0 & 0.0 & 0.0 \\
    \dots unmarked adjunct control (1) & 0.0 & 0.0 & 0.0
     \\
  \cline{1-4}
    \end{tabular}
}
    \caption{Detailed results for relation alignments and reentrancy alignments.}
    \label{tab:relationresults}
\end{table}

\topic{Qualitative Analysis} A number of errors from our subgraph aligner resulted from unseen multiword expressions in our test data that our span preprocessing failed to recognize and our aligner failed to align. For example, the expression ``on the one hand'' appears in test and should be aligned to \amr{contrast-01}. The JAMR aligner suffers without a locality bias; 
we notice several cases where it misaligns words that are repeated in the sentence.
%to resolve ambiguities in alignment when distance would be a useful feature, compared to our aligner. 
%There are several examples in test where simple words like \w{people}, \w{ant}, or \w{can} are repeated in a sentence and JAMR fails to align the correct token and concept. 
%The ISI aligner suffers from lower coverage and a willingness to align disconnected portions of the graph to the same span. 
The ISI aligner generally does not align very frequent nodes such as \amr{person}, \amr{thing}, \amr{country}, or \amr{name}, resulting in generally lower coverage. It also frequently aligns disconnected nodes with the same concept to one token instead of separate tokens. %, for example aligning \w{ant} $\rightarrow$ \amr{a1/ant}, \amr{a2/ant} instead of \w{ants} $\rightarrow$ \amr{a1/ant} and \w{ant} $\rightarrow$ \amr{a2/ant}. 
While our relation aligner yields significantly higher coverage, we do observe that the model is overeager to align relations to extremely frequent prepositions (such as \w{to} and \w{of}), resulting in lower precision of single relations in particular.

% JAMR: needs bolt12_10510_9811.12, people bolt12_10510_9811.12, bolt12_10510_9796.1 can, bolt12_10489_6271.8 dozens

% ISI: bolt12_10489_6271.8 call, bolt12_10510_8171.10 lives, bolt12_10510_8171.10 really, bolt12_10510_9796.1 ants, bolt12_10510_9811.5 general

% Ours: bolt12_2981_0166.1 have no choice, bolt12_3988_7641.18 On the one hand, bolt12_10510_9811.12 keep that in mind

\topic{Ablations} \Cref{tab:ablation} shows that projection distance is valuable, adding 1.20 points (exact align F1) for subgraph alignment\hide{\nss{does this include duplicates? I would guess it to be especially important for/with duplicates}} and 0.57 points for relation alignment. Despite showing anecdotal benefits in early experiments, the inductive bias does not aid the model in a statistically significant way. 
Using gold subgraphs for relation alignment produces an improvement of over 5 points, indicating the scope of error propagation for the relation aligner.

\begin{table}[t]
    \centering\small
    \begin{tabular}{|l|ccc|HHH}
    %\cline{2-7}
    \cline{2-4}
        \multicolumn{1}{l|}{\textbf{Ablations}} & \multicolumn{3}{c|}{\textbf{Exact Align}} \\ %& \multicolumn{3}{H|}{\textbf{Partial Align}}\\
        \multicolumn{1}{c|}{}  & P & R & F1 & P & R & F1\\
    %\cline{2-7}
    \cline{2-4}
    %\multicolumn{1}{c}{} & \multicolumn{6}{c}{Relation Alignments ($N=1345$)}\\
    \cline{1-4}
    Subgraphs  & 93.91 & 94.02 & 93.97 & 95.69 & 95.81 & 95.75\\
    Subgraphs ($-$distance)  & 92.69 & 92.85 & 92.77 & 93.88	& 94.32 & 94.1\\
    Subgraphs ($-$inductive bias) & 93.88 & 93.44 & 93.66 & 95.69 & 95.74 & 95.72 \\
    \cline{1-4}
     Relations & 85.67 & 85.37 & 85.52 & 88.74 & 88.44 & 88.59 \\
    Relations ($-$distance) & 85.14 & 84.77 & 84.95 & 88.14 & 87.76 & 87.95 \\
    Relations (gold subgraphs) & 91.21 & 90.59 & 90.90 & 93.00 & 92.37 & 92.68 \\
  \cline{1-4}
    \end{tabular}
    \caption{Results when the aligner is trained without projection distance probabilities ($-$distance) and without the subgraph inductive bias ($-$inductive bias), as well as a relation aligner with access to gold (instead of trained) subgraphs.}
    \label{tab:ablation}
\end{table}

\section{Conclusions}

We demonstrate structure-aware AMR aligners that combine the best parts of rule-based and statistical methods for AMR alignment. We improve on previous systems in terms of accuracy and particularly in terms of alignment coverage and variety of AMR components to be aligned. 

\section*{Acknowledgments}

We thank reviewers for their thoughtful feedback, Jakob Prange for assisting with annotation, and members of the NERT lab for their support.

\bibliography{amralign}
\bibliographystyle{acl_natbib}

\newpage
\appendix

\begin{table*}[h!]
    \centering\small
    \begin{tabular}{|l|ccc|ccc|c|}
    \cline{2-8}
        \multicolumn{1}{l|}{\textbf{IAA}} & \multicolumn{3}{c|}{\textbf{Exact Align}} & \multicolumn{3}{c|}{\textbf{Partial Align}} & \textbf{Spans}\\
        \multicolumn{1}{c|}{}  & P & R & F1 & P & R & F1 & F1\\
    \cline{2-8}
    %\multicolumn{1}{c}{} & \multicolumn{6}{c}{Relation Alignments ($N=1345$)}\\
    \hline
    Subgraphs (366)  & 94.54 & 94.54 & 94.54 & 95.56 & 95.56 & 95.56 & 94.97 \\[5pt]
    Relations (260) &  91.09 & 90.38 & 90.73 & 93.38 & 92.66 & 93.02 & 93.75 \\[5pt]
    Reentrancies (65) & 76.92 & 76.92 & 76.92 & 90.00 & 90.00 & 90.00 & 90.77 \\[5pt]
    Duplicates (5) & 75.00 & 60.00 & 66.67 & 79.17 & 63.33 & 70.37 & 66.67 \\[5pt]
  \hline
    \end{tabular}
    \caption{Interannotator Agreement for \textbf{subgraph}, \textbf{relation}, \textbf{reentrancy}, and \textbf{duplicate subgraph layers} of alignment scored on a sample of 40 sentences of the gold test data.}
    \label{tab:IAA}
\end{table*}

\section{Interannotator Agreement}\label{sec:IAA}

\Cref{tab:IAA} illustrates interannotator agreement for each of the four alignment layers.

\section{Identifying Spans}\label{sec:appendix-spans}
As a preprocessing step, sentences have their tokens grouped into spans based on three criteria, outlined in detail below:

\begin{enumerate}
    \item Named entity spans identified by Stanza.
    \item Spans matching multiword expressions from a fixed list of $\approx$1600
    \begin{enumerate}[leftmargin=0pt,itemindent=26pt]
        \item 143 prepositional MWEs from STREUSLE \citep{streusle,schneider-18}
        \item 348 verbal MWEs from STREUSLE
        \item 1095 MWEs taken from gold AMRs in LDC train data (any concept which is a hyphenated compound of multiple words, e.g., \textit{alma-mater} or \textit{white-collar}) and are not present in the above lists.
        \item $\approx$12 hand-added MWEs 
    \end{enumerate}
    \item Any sequence of tokens which is an exact match to a name in the gold AMR (e.g., ``United Kingdom'' and \amr{(n/name :op1 "United" :op2 "Kingdom")}) is also treated as a span.
\end{enumerate}

%\newpage

\section{Rule-based Subgraph Alignment Preprocessing}\label{sec:subgraph-preproc}

\subsection{Token matching}

We use three phases of rule-based alignment which attempt to align particular spans to particular AMR subgraphs:

\begin{enumerate}
    \item Exact token matching: If there is a unique full string correspondence between a span and a name or number in the AMR, they are aligned.
    %Spans such that there is an exact token match to a named entity or number in the gold AMR are aligned, assuming that only one span and one subgraph fit the exact match. 
    \item Exact lemma matching: If there is a unique correspondence between an AMR concept and the lemma of a span (which in the case of a multiword span is the sequence of lemmas of the tokens joined by hyphens), they are aligned.
    %Spans such that there is an exact lemma match to a concept in the gold AMR are aligned, assuming that only one span and one concept fit the exact match. 
    \item Prefix token matching: A span with a prefix match of length 6, 5, or 4 is aligned if it uniquely corresponds to an AMR named entity.
    \item Prefix lemma matching: A span with a prefix match of length 6, 5, or 4 of its lemma is aligned if it uniquely corresponds to an concept.
    \item English rules: Several hand-written rules for matching English strings to specific subgraphs are used to match constructions such as dates, currency, and some frequent AMR concepts with many different ways of being expressed, such as \amr{and} and \amr{-}.
    \begin{itemize}
        \item Parsing dates and times
        \item Numbers written out (e.g., \w{one}, \w{two}, \w{thousand}, etc.)
        \item Currencies (e.g., \w{\$}, \w{€}, etc.)
        \item Decades (e.g., \w{twenties}, \w{nineties})
        \item \amr{and} (matching \w{and}, \w{additionally}, \w{as well}, etc.)
        \item \amr{multi-sentence} (matching punctuation)
        \item \amr{:polarity -} (matching \w{not}, \w{none}, \w{never}, etc.)
        \item \amr{cause-01} (matching \w{thus}, \w{since}, \w{because}, etc.)
        \item \amr{amr-unknown} (matching \w{?}, \w{who}, \w{when}, etc.)
        \item \amr{person} (matching \w{people})
        \item \amr{rate-entity-91} (matching \w{daily}, \w{weekly}, etc.)
        \item \amr{"United" "States"} (matching \w{US}, \w{U.S.}, \w{American}, etc.)
        \item \amr{include-91} (matching \w{out of}, \w{include}, etc.)
        \item \amr{instead-of-91} (matching \w{instead}, etc.)
        \item \amr{have-03} (matching \w{have}, \w{'s}, etc.)
        \item \amr{mean-01} (matching \w{:} and \w{,})
        \item \w{how} (matching \amr{:manner thing} or \amr{:degree so})
        \item \w{as\dots as } (matching \amr{equal})
    \end{itemize}
    
\end{enumerate}

            % # exact match for 'how'
            % # as ... as construction

\subsection{Graph rules}
We also perform preprocessing to expand a subgraph alignment to include some neighboring nodes. These fall into two main categories:

\begin{enumerate}
    \item Some AMR concepts are primarily notational rather than linguistic and should be aligned together with a neighboring node. For example named entities (e.g., \amr{(country :name (n/name :op1 :United" :op2 "Kingdom"))}) are aligned as a unit rather than one node at a time. Likewise, date entities, and subgraphs matching \amr{(x/X-quantity :unit X :quant X)} or \amr{(x/X-entity :value X)} are also aligned as a unit.
    \item Neighboring nodes which are associated with morphological information of the aligned span (e.g., \w{biggest} $\rightarrow$ \amr{(have-degree-91 :ARG1 big :ARG2 most)}) are added to the alignment using a series of rules for identifying comparatives, superlatives, polarity, and suffixes such as \w{-er} or \w{-able}, etc.
\end{enumerate}

\section{Rule-based Relation Alignment Preprocessing}\label{sec:rel-preproc}

Many of the relations are forced to be aligned in a particular way as a matter of convention. We use a similar approach to that of \cite{groschwitz2018amr}.

\begin{enumerate}
    \item \amr{:ARGX} edges are automatically aligned to the same span as the parent (\amr{:ARGX-of} edges are automatically aligned to the child).
    \item \amr{:opX} edges are automatically aligned with the parent.
    \item \amr{:sntX} edges are automatically aligned with the parent.
    \item \amr{:domain} edges are automatically aligned with the parent. (We don't align these edges to copula. Instead, a concept with a \amr{:domain} edge is thought of as a predicate which takes one argument.)
    \item \amr{:name}, \amr{:polarity}, and \amr{:li} edges are automatically aligned with the child.
\end{enumerate}

\subsection{Token matching}
Some relations take the form \amr{:prep-X} or \amr{:conj-X} where \amr{X} is a preposition or conjunction in the sentence. We use exact match to align these relations as a preprocessing step. The relations \amr{:poss} and \amr{:part} may be automatically aligned to \w{'s} or \w{of} if the correspondence is unique within a sentence.

\section{Rule-based Reentrancy Alignment Preprocessing}\label{sec:reent-preproc}
Primary edges are identified as a preprocessing step before aligning reentrancies with the following rules: Any relation which is aligned to the same span as its token (any incoming edge which is a part of a span's argument structure) is automatically made the primary edge. Otherwise, for each edge pointing to a node, we identify the spans aligned to the parent and child nodes in the subgraph layer. Whichever edge has the shortest distance between the span aligned to the parent and the span aligned to the child is identified as the primary edge. In the event of a tie, the edge whose parent is aligned to the leftmost span is identified as the primary edge. Primary reentrancy edges are always aligned to the same span the edge is aligned to in the relation layer of alignments.

\end{document}